%% file: main.tex
\pdfoutput=1

\documentclass[11pt]{article}

\usepackage{acl}
\usepackage{booktabs}
\usepackage{multirow}

\usepackage{times}
\usepackage{latexsym}

\usepackage[T1]{fontenc}
\usepackage{makecell}

\usepackage[utf8]{inputenc}

\usepackage{microtype}

\usepackage{graphicx}
\usepackage{xcolor}
\usepackage{microtype}
\usepackage{enumitem}
\usepackage{algorithm}
\usepackage{algorithmic}
\usepackage{amsmath}
\usepackage{multirow}
\usepackage{booktabs}
\usepackage{url}
\usepackage{graphicx}
\definecolor{RoseQuartzBg}{HTML}{F7CAC9}
\definecolor{RoseQuartz}{HTML}{F5A798}
\definecolor{Serenity}{HTML}{92A8D1}
\definecolor{OrangeRed}{rgb}{1.0, 0.27, 0.0}
\definecolor{Red}{rgb}{1.0, 0.0, 0.0}
\definecolor{Turquoise}{HTML}{0F4C81}
\usepackage{xparse}
\usepackage{soul}
\usepackage{xspace}
\NewDocumentCommand{\lifu}{ mO{} }{\textcolor{OrangeRed}{\textsuperscript{\textit{Lifu}}\textsf{\textbf{\small[#1]}}}}
\NewDocumentCommand{\zoe}{ mO{} }{\textcolor{purple}{\textsuperscript{\textit{Zoe}}\textsf{\textbf{\small[#1]}}}}
\NewDocumentCommand{\sijia}{ mO{} }{\textcolor{blue}{\textsuperscript{\textit{Sijia}}\textsf{\textbf{\small[#1]}}}}

\title{A Survey of Document-Level Information Extraction }

\author{First Author \\
  Zoe / Address line 1 \\
  Affiliation / Address line 2 \\
  Affiliation / Address line 3 \\
  \texttt{zoez@vt.edu} \\\And
  Second Author \\
  Affiliation / Address line 1 \\
  Affiliation / Address line 2 \\
  Affiliation / Address line 3 \\
  \texttt{email@domain} \\}

\author{Hanwen Zheng \quad Sijia Wang \quad Lifu Huang \\
   Virginia Tech \\
 \texttt{\{zoez,sijiawang,lifuh\}@vt.edu}  \\}

\begin{document}
\maketitle
\begin{abstract}
Document-level information extraction (IE) is a crucial task in natural language processing (NLP). This paper conducts a systematic review of recent document-level IE literature. In addition, we conduct a thorough error analysis with current state-of-the-art algorithms and identify their limitations as well as the remaining challenges for the task of document-level IE. According to our findings, labeling noises, entity coreference resolution, and lack of reasoning,  severely affect the performance of document-level IE. The objective of this survey paper is to provide more insights and help NLP researchers to further enhance document-level IE performance.
\end{abstract}

\input{1_introduction}
\input{2_task}

\input{4_datasets}

\input{3_evaluation}

\input{5_methods}

\input{6_discussion}
\input{7_challenges}





\bibliography{custom}
\bibliographystyle{acl_natbib}

\appendix
\input{appendix}

\end{document}

%% file: 1_introduction.tex
\section{Introduction}
\label{sec:intro}
Natural language processing (NLP) triggers the present wave of artificial intelligence \citep{vaswani2017attention, dosovitskiy2021an, 9710580, Zhang2021AttentionbasedNN, Zhang2022AQI}. Information Extraction (IE) plays a vital role in all aspects of NLP by extracting structured information from unstructured texts \cite{lin-etal-2020-joint, wang-etal-2022-query}. Document-level IE has witnessed significant progress, benefiting from the enormous data resources provided by the Internet and the rapidly growing computational power resources~\cite{yao_docred_2019, xu2021documentlevel, tong_docee_2022}. However, several challenges persist within the realm of document-level IE research, such as entity coreference resolution, reasoning across long-span contexts, and a lack of commonsense reasoning. Furthermore, current document-level IE research predominantly focuses on restricted domains and languages~\cite{zheng_doc2edag_2019, yang_dcfee_2018, tong_docee_2022, li_document-level_2021}, which results in challenges for model comparisons and generalization. This limitation poses difficulties in conducting model comparisons and hampers the generalizability of findings.

To fulfill the aforementioned challenges, this survey reviews recent document-level relation extraction \textbf{(doc-RE)} and document-level event extraction \textbf{(doc-EE)} models and datasets to inform and encourage researchers for multilingual and cross-domain studies.  In addition, we conduct a thorough error analysis among existing models and discuss these errors. Finally, we summarize the current literature work and propose potential future improvements to document-level IE research. The contributions of this survey paper include:
\begin{itemize}
    \item{We systematically summarize and categorize the existing datasets and approaches for Doc-RE and Doc-EE.}
    \item{A thorough error analysis is conducted with current state-of-the-art (SOTA) algorithms.}
    \item{To identify the current model challenges and limitations, we analyze and discuss the errors and construct error statistics.}
\end{itemize}

This survey aims to contribute to the NLP community by providing valuable insights into document-level IE tasks. Our analysis of errors encountered in this study will serve as a foundation for future advancements in document-level IE research, encouraging researchers to innovate and improve upon existing methodologies. It is our hope that these findings will contribute to a deeper understanding of document-level IE and stimulate further enhancements in this field of study.


%% file: 2_task.tex
\section{Tasks Definition}
\label{sec:task definition}
\begin{figure*}[ht]
    \centering
    \includegraphics[width=0.9\linewidth]{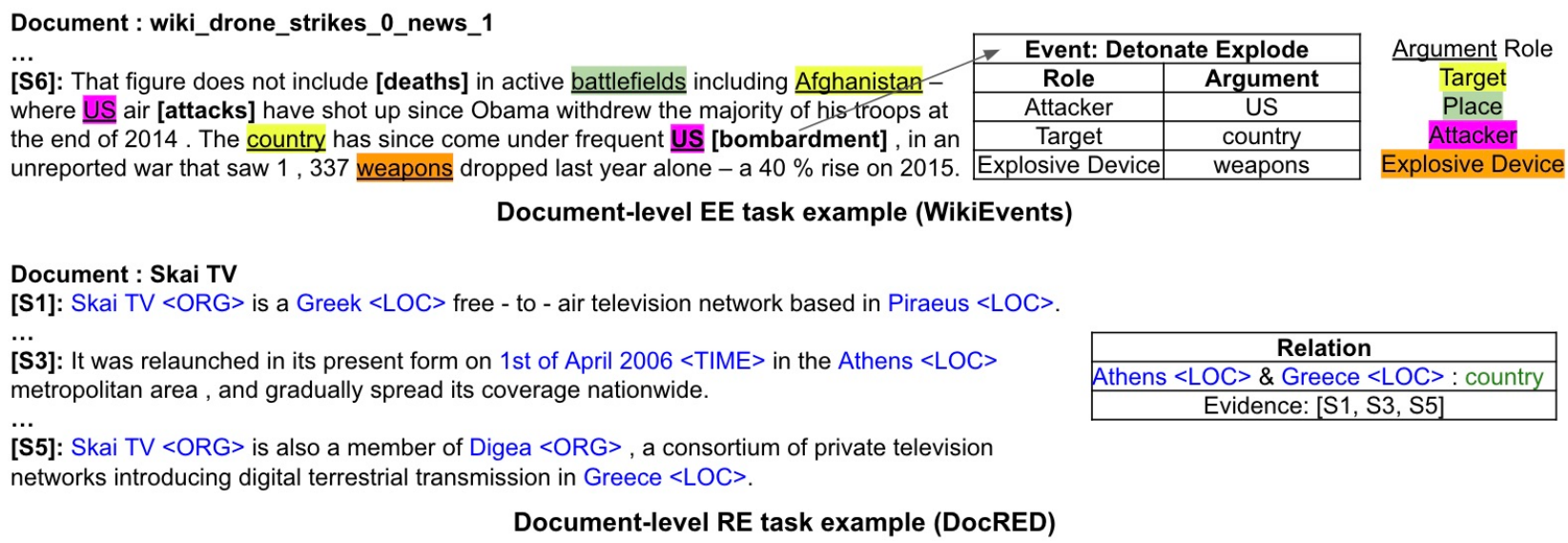}
    \caption{Examples of doc-EE and doc-RE.
    }
    \label{fig:task_example}
\end{figure*}
\subsection{Event Extraction}
{Event extraction~\cite{grishman1997information,chinchor1998muc,ahn2006stages} is a task to identify and classify event triggers and relevant participants from natural language text. }
{Formally, }
given a document consisting of a set of sentences where each sentence consists of a sequence of words,  
the objective of this task is to identify and extract the following components from a given document: 
\textbf{Event Mention}, which refers to the phrases or sentences denoting an event; 
\textbf{Event Trigger}, typically in the form of a verb that signals the occurrence of an event; 
\textbf{Event Type}, indicating the predefined type of event specified by the dataset, such as Conflict-Attack; 
\textbf{Argument Mention}, comprising entity mentions that provide additional details on the event, such as who, what, when, where, and how the event occurred; 
\textbf{Argument Role}, representing the role or type of argument associated with the entity;
and finally, \textbf{Event Record}, the entry in an event table, containing several arguments with argument roles.

\subsection{Relation Extraction}
The task of Relation Extraction involves predicting attributes and relationships between entities mentioned in a given document {\cite{zhou2005exploring}}.
Given a document $D$ with a set of sentences, we assume that 
$D$ also contains a set of entities $V=\{e_i\}_{i=1}^N$. 
For each entity $e_i$, it might contain multiple entity mentions $e_i=\{m_j\}_{j=1}^M$. The doc-RE task is to predict the relation types between an entity pair $(e_s,e_o)_{s,o\in\{1,\cdots, N\},s\ne o}$, where $s$ stands for the subject and $o$ stands for the object. 
It is possible for an entity pair to have multiple relations that require prediction, thereby rendering the task a multi-label classification problem.


More specifically,  \textbf{Entity} refers to units such as \textit{People}, \textit{Geographic Entity}, \textit{Location}, \textit{Organization}, \textit{Date}, and \textit{Number} within a text. \textbf{Entity Mention} refers to a phrase within a text that identifies a specific entity. For instance, ``NYC'' and ``the big apple'' are both entity mentions for ``New York City''. \textbf{Intra-sentence Relation} describes the relationship between entities within a single sentence, and the features within are often referred to as local features. On the other hand, \textbf{Inter-sentence Relation} refers to the relationship between entities across multiple sentences, and the features within are often referred to as global features.

%% file: 4_datasets.tex
\section{Datasets}
\label{sec:datasets}

Existing studies only evaluate their proposed approaches on restricted targeted domains or languages. As a result, it is challenging to compare the effectiveness of different methods under a more general scenario. In this section, we list all doc-EE and doc-RE datasets, to share all possible options with the research community. 

\subsection{Doc-RE Datasets}

\begin{table*}
\small \centering
\begin{tabular}{lllllll}
\toprule
\textbf{Dataset} & \textbf{Annotation} & \textbf{\# Rel Types} & \textbf{\# Rel Facts} & \textbf{\# Train} &\textbf{\# Dev}  & \textbf{\#Test}\\
\midrule
DGM \cite{jia-etal-2019-document}  &Distant Supervision &1  & -& 32,040& -&-\\
CDR \cite{luan-etal-2018-multi}  &Human-annotated &1 & - & 1,500 & 500&500\\
GDA \cite{Wu2019RENETAD}  &Distant Supervision &1  & -& 30,192& 5,839&1,000\\
DocRED \citep{yao2019DocRED} & Distant Supervision & 96  & 50,345&3,053&1,000 &1,000\\
Re-DocRED \citep{tan_revisiting_2022} & Combined & 96  &120,664& 3,053& 500 &500\\
SciREX \citep{jain-etal-2020-scirex}  & Human-annotated & 2 & - & 438 & 131&131\\
HacRED \cite{Cheng2021HacREDAL}  &Combined &26  & 65,225& 9,231 &1,500&  1,500\\
\bottomrule
\end{tabular}
\caption{
Statistics of Doc-RE datasets.
}
\label{tab:re_datasets}
\end{table*}

For biomedical domain, \textbf{Drug-gene-mutation (DGM)} \cite{jia-etal-2019-document} contains 4,606 PubMed articles, which are automatically labeled via distant supervision. DGM annotations include three entity types: \textit{drugs}, \textit{genes}, and \textit{mutations}, and three relation types, including \textit{drug-gene-mutation}, \textit{drug-mutation}, and \textit{gene-mutation relations}. 
\textbf{GDA} \cite{Wu2019RENETAD} gene-disease association corpus contains 30,192 titles and abstracts from PubMed articles that have been automatically labeled for \textit{genes}, \textit{diseases}, and \textit{gene-disease associations} via distant supervision. \textbf{CDR} \cite{luan-etal-2018-multi} is manually annotated for \textit{chemicals}, \textit{diseases}, and \textit{chemical-induced disease (CID)} relations by domain experts. It contains the titles and abstracts of 1,500 PubMed articles and is split into training, validation, and test sets equally. 


Several Doc-RE datasets are constructed for other domains or languages. \textbf{DocRED} \cite{yao2019DocRED} is a human-annotated doc-RE dataset, that includes 132,375 entities and 56,354 relational facts annotated on 5,053 Wikipedia documents. Doc-RED is generated by mapping Wikidata triples, originating from a comprehensive knowledge base closely intertwined with Wikipedia, onto complete English Wikipedia documents to get entity annotations. \textbf{RE-DocRED} \cite{tan_revisiting_2022} refines 4,053 documents in the DocRED dataset targeting on resolving the problem of false negative samples. RE-DocRED increased the relation triples from 50,503 to 120,664 and decreased the \textit{no\_relation} samples by $3.1\%$ by adding the missing relation triples back to the original DocRED.
\textbf{SciREX} \cite{jain-etal-2020-scirex} is a document-level IE dataset that contains multiple IE tasks. It mainly focuses on its doc-RE tasks, such as Binary and N-ary relation classification. It consists of both automatic and human-annotated articles in the computer science field. \textbf{HacRED} \cite{Cheng2021HacREDAL} is a Chinese doc-RE dataset collected from CN-DBpedia \cite{Xu2017CNDBpediaAN} that focuses on hard cases, such as long text and long distance between argument pairs, containing distractors or multiple homogeneous entity mentions. 

\subsection{Doc-EE Datasets}
Doc-EE datasets are mainly collected from the news and financial domain. News is a large-scale accessible source of events like social emergencies and human life incidents, thus many datasets are created focusing on news events. Meanwhile, exploding volumes of digital financial documents, as a byproduct of continuous economic growth, have been created. Many datasets are created to help extract valuable structured information to detect financial risks or profitable opportunities.
Statistics of the datasets for Doc-EE are summarized in Table \ref{tab: EE stats}. 
\begin{table*}
\centering
\small
\begin{tabular}{lllllll}
\toprule
\textbf{Dataset} & \textbf{\# Docs} & \textbf{\# Events} & \textbf{\# Event types} & \textbf{\# Roles} & \textbf{\# Arguments} & \textbf{Ratio}\\
\midrule
ACE-2005\footnote{\url{https://catalog.ldc.upenn.edu/LDC2006T06}}  & 599 & 4,202 & 33 & 35& 9,590& -\\
MUC-4 \cite{muc-1992-message} & 1,700 & 1,514 & 4 & 5& 2,641& 13:2:2\\
RAMS \cite{ebner-etal-2020-multi} & 9,124 & 8,823 & 139 & 65 & 21,237 &  8:1:1\\
WikiEvents \citep{li_document-level_2021} & 246 & 3,951 & 50 &  59 & 5,536 & 10:1:1\\
DocEE \cite{tong_docee_2022} & 27,485 & 27,485 & 59 & 356 & 180,520 & -\\
\midrule
ChFinAnn \citep{zheng2019doc2edag} & 32,040 & 47,824 & 5 & 35& 289,871 & 8:1:1\\
DCFEE \citep{yang_dcfee_2018} & 2,976 & 3,044 & 4 & 35& - & 8:1:1\\
DuEE-Fin \citep{zheng2019doc2edag} & 11,699 & 15,850 & 13 & 92& 81,632 & 6:1:3\\
\bottomrule
\end{tabular}
\caption{Statistics of Doc-EE datasets.}
\label{tab: EE stats}
\end{table*}

For the news domain, \textbf{ACE-2005\footnote{\url{https://catalog.ldc.upenn.edu/LDC2006T06}}} is a sentence-level event extraction (SEE) \cite{wang-etal-2022-query, wang-etal-2022-prompt} dataset but has been frequently used for comparison in doc-EE. 
Unlike ACE-2005 which contains 5 groups of events covering \textit{justice}, \textit{life}, \textit{business events}, etc, \textbf{MUC-4} \cite{muc-1992-message} focuses on one specific event type, \textit{attack} events. MUC-4 contains 1,700 human-annotated news reports of terrorist attacks in Latin America collected by Federal Broadcast Information Services. More specifically, MUC-4 includes six incident types: \textit{attack}, \textit{kidnapping}, \textit{bombing}, \textit{arson}, \textit{robbery}, and \textit{forced work stoppage}, and four argument roles, including \textit{individual perpetrator}, \textit{organization perpetrator}, \textit{physical target}, and \textit{human target}.
\textbf{WikiEvents} \citep{li_document-level_2021} follows the ontology from the KAIROS project\footnote{\url{https://www.ldc.upenn.edu/collaborations/currentprojects}} for event annotation, which defines 67 event types in a three-level hierarchy.  Researchers used the BRAT interface for online annotation of event mentions (triggers and arguments) and event coreference separately. 
\textbf{Roles Across Multiple Sentences ({RAMS})} \cite{ebner-etal-2020-multi} is a crowd-sourced dataset with 9,124 event annotations on news articles from Reddit following the AIDA ontology. 
\textbf{DocEE} is the largest Doc-EE dataset to date. DocEE uses historical events and timeline events from Wikipedia as the candidate source to define 59 event types and 356 event argument roles. This dataset includes 27,485 document-level events and 180,528 event arguments that are manually labeled. 




For the financial domain, \textbf{DCFEE} \citep{yang_dcfee_2018} comes from companies' official finance announcements and focuses on four event types: \textit{Equity Freeze}, \textit{Equity Pledge}, \textit{Equity Repurchase}, and \textit{Equity Overweight}. Data labeling was done through distant supervision.
\textbf{ChFinAnn} \citep{zheng2019doc2edag} contains official disclosures such as annual reports and earnings estimates, obtained from the Chinese Financial Announcement (CFA). The dataset has five event
types: \textit{Equity Freeze}, \textit{Equity Repurchase},
\textit{Equity Underweight}, \textit{Equity Overweight}
and \textit{Equity Pledge}, with 35 different
argument roles in total. In contrast to Doc-EE with one event in each document,  29.0\% of the documents in ChFinAnn contain multiple events. 
\textbf{DuEE-Fin} \citep{zheng2019doc2edag} is the largest human-labeled Chinese financial dataset. It is collected from real-world Chinese financial news and annotated with 13 event types. 29.2\% of the documents contain multiple events and 16.8\% of events consist of multiple arguments. 


%% file: 3_evaluation.tex
\section{Evaluation Metrics}
\label{sec:evaluation}

In document-level information extraction (IE), the primary evaluation metrics are Precision (P), Recall (R), and Macro-F1 score \cite{kowsari2019text}. Additionally, for doc-RE, Ign F1 is used as an evaluation metric \citep{yao2019DocRED}. Ign F1 refers to the F1 score that excludes relational facts shared by the training and dev/test sets. This metric is important for evaluating the generalizability of the model, as it disregards triples that are already present in the annotated training dataset. 




\begin{table*}[ht]
\centering\small\scalebox{0.85}{
\begin{tabular}{p{1.2cm}p{3.0cm}p{2.8cm}p{9cm}}
\toprule
\textbf{Task} & \textbf{Main Category} & \textbf{Sub Category} & \textbf{Approaches}\\
\midrule

\multirow{16}{*}{Doc-RE} 
                        & \multirow{5}{*}{Multi-granularity-based} &  Sentence-level$\to$ Paragraph-level$\to$ Document-level&\citet{tang_hin_2020}\\ \cmidrule{3-4}
                        &                                  & Mention-level$\to$ Entity-level & \citet{jia_document-level_2019} \\ 
                        \cmidrule{2-4}
                        & \multirow{5}{*}{Graph-based} & \multirow{4}{*}{Heterogeneous graph }&\citet{quirk_distant_2017}, \citet{peng_cross-sentence_2017}, \citet{song_n-ary_2018}, \citet{guo_attention_2019}, \citet{sahu_inter-sentence_2019}, \citet{christopoulou_connecting_2019}, \citet{wang_global--local_2020}, \citet{xu_document-level_2021}, \citet{zeng_double_2020}, \citet{li_graph_2020}, \citet{zhang_document-level_2020}, \citet{xu_document-level_2023}, \citet{xu_discriminative_2021}
  \\ \cmidrule{3-4}
                        &                            &  Homogeneous graph&\citet{nan_reasoning_2020}
\\ 
                        \cmidrule{2-4}
                        & \multirow{2}{*}{Sequence-based} & Neural Networks & \citet{xu_entity_2021}, \citet{zhang_document-level_2021} \\ \cmidrule{3-4}
                        &                               & Attention\textbackslash Transformer & \citet{zhou_document-level_2021}, \citet{tan_document-level_2022}
 \\ \cmidrule{2-4}
 & \multirow{2}{*}{Evidence-based} &  Path reasoning & \citet{huang_three_2021}\\ \cmidrule{3-4}
                        &                                 & Evidence retrieval&\citet{xie_eider_2022}, \citet{xiao_sais_2022}
 \\ 
                        
\midrule
\multirow{10}{*}{Doc-EE} & \multirow{3}{*}{Multi-granularity-based} &  Sentence-level$\to$ Paragraph-level$\to$ Document-level&\citet{yang_dcfee_2018}, \citet{huang_exploring_2021}\\ 
                        \cmidrule{2-4}
                        & \multirow{1}{*}{Graph-based} & \multirow{1}{*}{Heterogeneous graph
                        } &\citet{zheng2019doc2edag}, \citet{xu_document-level_2021}, \citet{zhu_efficient_2022}, \citet{xu_two-stream_2022}
  \\ \cmidrule{2-4}
                        & \multirow{2}{*}{Sequence-based} & Neural Networks & \citet{huang_document-level_2021-1} \\ \cmidrule{3-4}
                        &                               & Attention\textbackslash Transformer & \citet{yang_document-level_2021}, \citet{liang_raat_2022}\\\cmidrule{2-4}
                         & \multirow{1}{*}{Generation-based} & -&\citet{li_document-level_2021}, \citet{zeng_ea2e_2022}
 \\\cmidrule{2-4}& \multirow{1}{*}{Memory-based} & - &\citet{du_dynamic_2022}, \citet{cui_document-level_2022}\\
                                                    
\bottomrule
\end{tabular}}
\caption{Typology of Doc-IE methods.}
\label{tab:Typology}
\end{table*}

%% file: 5_methods.tex

\section{Methods}
\label{sec:method}
The fundamental challenge in doc-RE and doc-EE is to express document content in a concise and effective way such that key information is maintained. Previous approaches usually resort to hierarchical, graph-based, or sequential structures. More recently, due to the emergence of powerful generative pre-trained language models (PLMs), generative models have also been introduced to address doc-IE tasks. A typology of existing doc-RE and doc-EE approaches categorized by model design is shown in Table \ref{tab:Typology}.

\subsection{Doc-RE Approaches}

\paragraph{Multi-granularity-based Models} The multi-granularity-based approach aims to emphasize the use of information from different granularities and the aggregation of global information. The standard procedure involves concatenating features from each level to complete the IE tasks. \citet{jia_document-level_2019}  approaches document-level N-ary relation extraction using a multiscale representation learning method. This approach aggregates the representations of mentions and ensembles multiple sub-relations.
The \textbf{HIN} (Hierarchical Inference Network) \cite{tang_hin_2020} uses Bi-LSTMs at the token, sentence, and document levels to extract features as sequences and weighs the overall features with the attention mechanism to obtain both local and global information.
Multi-granularity-based designs employ two strategies: either they address intermediate tasks using various models, or they utilize the same model in a hierarchically ordered manner to independently tackle each subtask of information extraction, such as from sentence level to document level.

\paragraph{Graph-based Models} Graph-based models generally construct a graph with words, mentions, entities, or sentences as nodes and define different types of edges across the entire document, further predicting the relations by reasoning on the graph. 
The first work done on doc-RE using a graph-based method is \textbf{DISCREX} \cite{quirk_distant_2017}, where a document graph is constructed with word nodes and edges representing intra- and inter-sentential relations including dependency, adjacency, and discourse relations. \citet{peng_cross-sentence_2017} contributes a Graph-LSTMs model with a bidirectional LSTM consisting of two directed acyclic graphs (DAG), and edges representing relations between nodes. \citet{song_n-ary_2018} further compares bidirectional graph LSTM with bidirectional DAG LSTM, finding that the former, which doesn't alter the input graph structure, exhibits superior performance. While such dependency graphs have rich structural information, the pruning strategy does not necessarily keep the relevant information. \textbf{AGGCNs} \cite{guo_attention_2019} proposes an end-to-end neural network that encodes the entire graph using multi-head self-attention to learn edge weights based on paired relations and using densely connected layers to glean global information.  \citet{sahu_inter-sentence_2019} designates words as individual nodes and establishes five types of edges to represent inter-and intra-sentence dependency. The model then uses an edge-oriented GCNN to retain aggregated node representation.

\textbf{EoG} \cite{christopoulou_connecting_2019} is a pioneering graph-based model. It uses entities as nodes and forms unique edge representations through the paths between nodes to better capture the paired relations. 
To predict relations between entity pairs, EoG makes iterative inferences on the path between the entities and aggregates every edge to a direct entity-entity (EE) edge. 
Many papers adapted from EoG can be divided into two main categories:  homogeneous and heterogeneous graphs. 
\textbf{LSR} \cite{nan_reasoning_2020} uses graph structure as a latent variable to form a homogeneous graph. Unlike EoG which uses a human-constructed graph, LSR learns structured attention to refine the graph dynamically and constructs latent structures based on the previous refinement. 
For heterogeneous graphs, different types of edges are defined, representing unique features, functions, and even dual graphs. 
\textbf{GLRE} \cite{wang_global--local_2020} utilizes a multi-layer R-GCN to learn entity global representations which are used as queries in the multi-headed self-attention layer to learn entity local representations while using sentence-level information as the keys. 
\textbf{HeterGSAN} \cite{xu_document-level_2021} is a heterogeneous graph based on EoG that uses a GAT to encode the graph relying more on related entity pairs' attention. 

Dual graphs are normally used to capture hierarchical information. \textbf{GAIN} \cite{zeng_double_2020} utilized a heterogeneous mention-level graph to model interactions between the document and all mentions. 
\textbf{GEDA} \cite{li_graph_2020} 
optimized entity representation with two attention layers and a heterogeneous GCN layer. 
\textbf{DHG} \cite{zhang_document-level_2020} contains two heterogeneous graphs: a structure modeling graph using words and sentences as nodes to better capture document structure information and a relation reasoning graph using mentions and entities as nodes to perform multi-hop relation reasoning.
\textbf{POR} \cite{xu_document-level_2023} 
is a path-retrieving method between pair entities based on the BFS algorithm, which extracts path features through an LSTM and combines them using the attention mechanism.
\textbf{DRN} \cite{xu_discriminative_2021} passes encoded sentence and entity as a heterogeneous graph to a multi-layer GCN and meanwhile, uses the self-attention mechanism to learn a more contextual document-level representation. 


\paragraph{Sequence-based Models} Sequence-based models mostly rely on NN-based or Transformer-based architectures, which can model complex interactions among entities by implicitly capturing long-distance dependencies. 
\textbf{SSAN} \cite{xu_entity_2021} integrates structural dependencies within and throughout the encoding stage of the network, not only enabling simultaneous context reasoning and structure reasoning but also efficiently modeling these dependencies in all network layers. \textbf{ATLOP} \cite{zhou_document-level_2021} simply applies BERT's own attention weights for Localized Context Pooling as well as a dynamic adaptive thresholding strategy, to ensure that each entity maintains the same representation and balances the logits of positive and negative labels.
\textbf{DocuNet} \cite{zhang_document-level_2021} divides model construction into three parts leveraging a u-shaped semantic segmentation network to refine entity feature extraction. 
\textbf{KD} \cite{tan_document-level_2022} calculates self-attention in the vertical and horizontal directions of an $n\!\times\!n$ two-hop attention entity pair table using axial attention. 
The logits of paired entity relations are ranked with the logits of the threshold classes individually instead of ranking all positive logits together. Sequence-based approaches focus on capturing contexts and entity information via careful designs, either an adequate neural network structure or a novel loss function.

\paragraph{Path(Evidence)-based Models} Path-based models construct evidence paths and make relational decisions by reasoning on crucial information between entity pairs or sentences, instead of extracting features from the complete document. \textbf{THREE} \cite{huang_three_2021} presents three kinds of paths to find the supporting sentences: consecutive paths, multi-hop paths, and default paths for entity pairs.
\textbf{EIDER} \cite{xie_eider_2022} defines ``evidence sentences'', as a minimal number of sentences needed to predict the relations between certain pairs of entities in a document. 
\textbf{SAIS} \cite{xiao_sais_2022} utilizes two intermediary phases to obtain evidence information: pooled evidence retrieval, which distinguishes entity pairs with and without supporting sentences, and fine-grained evidence retrieval, which produces more interpretable evidence specific to each relation of an entity pair.
Those papers typically utilize supporting sentences from the DocRED dataset. When humans perform relation extraction on the long span of texts, we read through the whole document and evaluate sentences that are important for the task. The path-based approach is consistent with human perception and intuition, which has shown extraordinary performance. 

\subsection{Doc-EE Approaches}

\paragraph{Multi-granularity-based Models} 
\textbf{DCFEE} \cite{yang_dcfee_2018} first designs a SEE component to obtain the event arguments and event trigger and splices them together to get the input for the second component-DEE. The DEE uses a convolutional neural network to concatenate the output of SEE and the vector representation of the current sentence. 
\textbf{SCDEE} \cite{huang_exploring_2021} uses Graph Attention Networks (GAT) to transform vertex features, which are used to detect sentence communities and then obtain event types at the sentence level.

\paragraph{Graph-based Models}
\textbf{Doc2EDAG} \cite{zheng2019doc2edag} first identifies all the entities in a document and uses transformer fusing information at the document level. When an event type is triggered, the model starts to generate an entity-based directed acyclic graph (EDAG) and treats the Doc-EE task as an event table-filling task. Following the order of roles in an event type, EDAG decides which entity node to be expanded and considers a path-expanding sub-task until the EDAG is fully recovered.
\textbf{GIT} \cite{xu_document-level_2021} designs a heterogeneous graph with four types of edges between sentences and mentions.
Based on detected event types, a tracker is designed to extract corresponding arguments by expanding a constrained event type tree while tracking and storing records in global memory. 
\textbf{PTPCG} \cite{zhu_efficient_2022} calculates the semantic similarity between entities to construct a pruned complete graph after event and argument detection. 
Pruning is done by deciding whether entity pairs retain an edge based on heuristics. 
\textbf{TSAR} \cite{xu_two-stream_2022} 
leverages an AMR-guided interaction module to generate both global and local contextualized representations. A gate function is designed to decide the portion of global and local representation, to predict the argument roles for potential spans. 

\paragraph{Sequence-based Models} \textbf{DE-PPN} \cite{yang_document-level_2021} is an encoder-decoder doc-EE model which utilizes two transformers to identify sentence-level elements as the document encoder and a multi-granularity decoder to decode event, role, and event-role in parallel. 
 \textbf{ReDEE} \cite{liang_raat_2022} is the first to use entity relation information for doc-EE tasks, which utilizes SSAN to extract relation triples and transfer them with entity and sentence dependency.
\textbf{DEED} \cite{huang_document-level_2021-1} is an end-to-end model that utilizes Deep Value Networks (DVN), a structured prediction algorithm that effectively bridges the disparity between ground truth and prediction. This model directly incorporates event trigger prediction into DVN, thereby efficiently capturing cross-event dependencies for document-level event extraction. 

\paragraph{Generative Models}
The generative models are commonly found in doc-EE and joint extraction. \textbf{Bart-Gen} \cite{li_document-level_2021} takes the document and event templates as input, and uses an encoder-decoder model to generate arguments to fill in the blank in the templates based on the previous word in the sentence. \textbf{EA2E} \cite{zeng_ea2e_2022} aims to achieve event-aware argument extraction by labeling arguments from nearby events in the document to enhance the context. 

\paragraph{Memory-based Models}
\citet{du_dynamic_2022} introduces a memory-enhanced neural generation-based framework based on a sequence-to-sequence PLM. The memory stores gold-standard events and previously generated events of the same document; and the decoder retrieves event knowledge and decodes arguments dynamically based on the event dependency constraints. \textbf{HRE} \cite{cui_document-level_2022} emulates the human reading process by conducting a two-stage analysis - rough reading and elaborate reading. The initial rough reading detects the event type and saves it as memory tensors. Upon detection, elaborate reading extracts the complete event record with arguments and stores them in memory while updating with previous event type and argument memory tensors.


%% file: 6_discussion.tex
\section{Discussion}

\begin{figure}[t]
    \centering
    \includegraphics[width=0.9\linewidth]{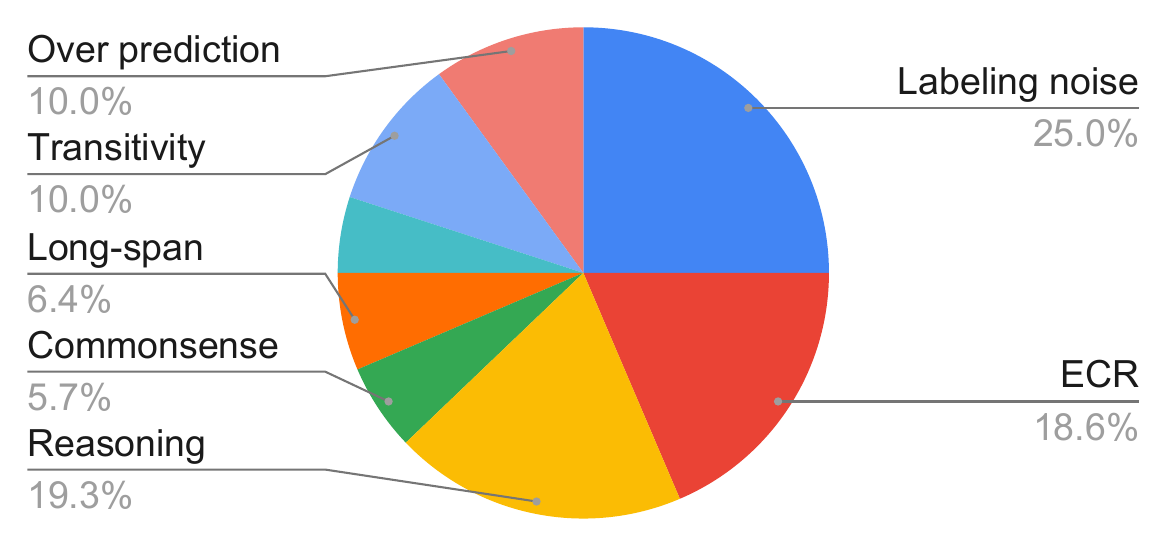}
    \caption{Doc-RE error distribution in DocRED and Re-DocRED
    }
    \label{fig:docRE-errors}
\end{figure}

\begin{figure}[t]
    \centering
    \includegraphics[width=0.9\linewidth]{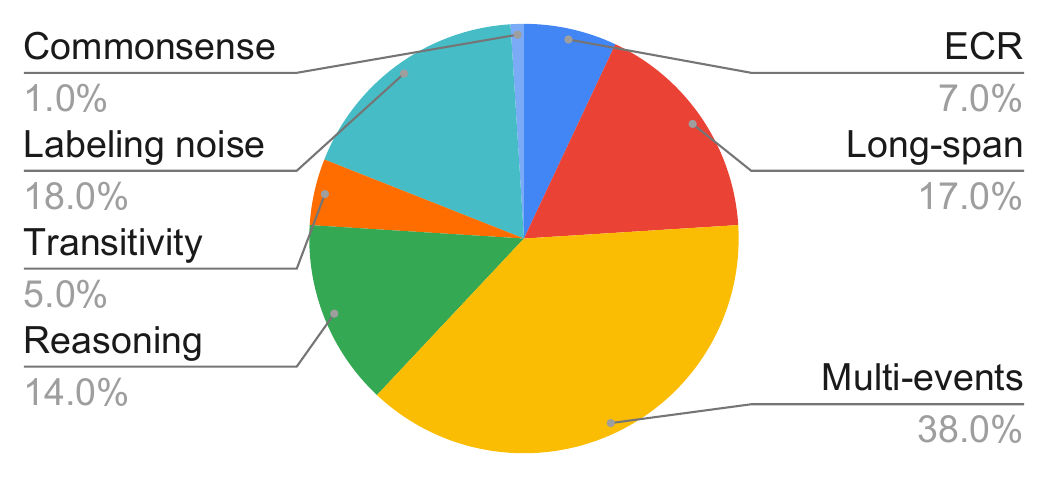}
    \caption{Doc-EE error distribution in ChFinAnn
    }
    \label{fig:ChFinAnn-errors}
\end{figure}

\begin{figure}[t]
    \centering
    \includegraphics[width=0.9\linewidth]{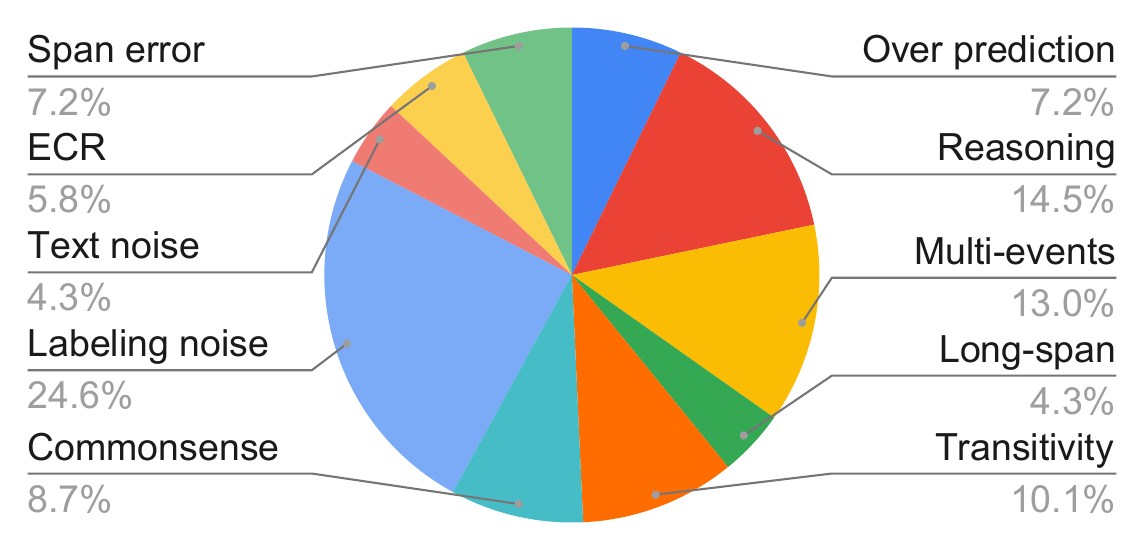}
    \caption{Doc-EE error distribution in WikiEvents 
    }
    \label{fig:WikiEvents-errors}
\end{figure}

We concluded seven major types of errors in three existing doc-RE works based on the DocRED and Re-DocRED datasets, as well as in four doc-EE works based on the WikiEvents and ChFinAnn datasets. Examples and distributions of each type are shown in Table~\ref{tab:corner_cases}, ~\ref{tab:re_errors}, and Figure~\ref{fig:docRE-errors}, ~\ref{fig:ChFinAnn-errors}, ~\ref{fig:WikiEvents-errors}. 

\paragraph{Entity coreference resolution} Document-level texts contain a large number of recognized entities along with coreferential words such as them, he, which, etc. Entity coreference resolution errors happen when the model fails to resolve all mentions in a document that refer to the same entity. 
\paragraph{Reasoning error} This type of error mainly relates to multi-hop logical reasoning. Document-level texts contain considerable amounts of information, so models may fail to give correct logical inferences based on the given information. Inferring from multi-hop information requires a model to have a high level of natural language understanding ability. 
\paragraph{Long-span} Document contains multiple sentences in a long span. This error happens when the model fails to capture the full context of a document or uses global information for inference. 
\paragraph{Commonsense knowledge} The error occurs when models fail to correctly extract relations or events or assume the wrong semantics due to a lack of commonsense and background knowledge, which humans are able to learn or understand instinctively. Many datasets are specific to some domains, and in the absence of relevant background and domain-specific knowledge models may inaccurately reason or misinterpret information. 
\paragraph{Relation transitivity error} Documents tend to have many entities appearing in the same sentence or across sentences. Relation transitivity errors occur when a model fails to correctly infer a relation between two entities based on their individual relations with a third entity. 
Additionally, not all relations are transitive, thus the model should correctly recognize when transitivity applies. 
\paragraph{Over prediction error} This error type refers to the spurious error (as we presented in Table 4) where there is no ground truth relation between two entities but the model predicts a relation, and can be caused by a number of reasons. For instance, when using large pre-trained language models to encode the documents, learned prior can cause models to make overconfident predictions. 



\begin{table*}[ht]
\small\centering\scalebox{0.8}{
\begin{tabular}{p{1.7cm}p{9cm}p{3.5cm}p{3.8cm}}
\toprule
 \textbf{Error Type} & \textbf{Text} & \textbf{GT} & \textbf{Prediction}  \\
\midrule
 Spurious error | Over prediction &\textbf{The Link River} $_{\mathrm{<LOC>}}$ is a short river connecting \textbf{Upper Klamath Lake} $_{\mathrm{<LOC>}}$ to \textbf{Lake Ewauna} $_{\mathrm{<LOC>}}$ in the city of \textbf{Klamath Falls} $_{\mathrm{<LOC>}}$ in the \textbf{U.S.}$_{\mathrm{<LOC>}}$ state of \textbf{Oregon} $_{\mathrm{<LOC>}}$. & N/A & Lake Ewauna $_{\mathrm{<LOC>}}$, Oregon $_{\mathrm{<LOC>}}$ : located in the administrative territorial entity\\
\midrule
 Spurious error | Learned prior &\textbf{Ngoako Ramatlhodi} $_{\mathrm{<PER>}}$, a senior member of the \textbf{African National Congress} $_{\mathrm{<ORG>}}$ , was \textbf{South Africa} $_{\mathrm{<LOC>}}$ 's Minister & N/A & African National Congress $_{\mathrm{<ORG>}}$, South Africa $_{\mathrm{<LOC>}}$ : country\\
\midrule
 Relation transitivity &At the \textbf{2007} $_{\mathrm{<TIME>}}$ \textbf{European Indoor Athletics Championships} $_{\mathrm{<MISC>}}$ he won a silver medal in the \textbf{4 x 400 metres} $_{\mathrm{<NUM>}}$ relay , with teammates Ivan \textbf{Buzolin} $_{\mathrm{<PER>}}$ , \textbf{Maksim Dyldin} $_{\mathrm{<PER>}}$ and \textbf{Artem Sergeyenkov} $_{\mathrm{<PER>}}$ &Artem Sergeyenkov $_{\mathrm{<PER>}}$, European Indoor Athletics Championships $_{\mathrm{<MISC>}}$ : participant of  & N/A\\

\bottomrule
\end{tabular}
}
\caption{
Corner case errors of RE models
}
\label{tab:corner_cases}
\end{table*}

\begin{table*}[ht]
\small\centering\scalebox{0.8}{
\begin{tabular}{p{2cm}p{10cm}p{4.5cm}p{0.8cm}}
\toprule
 \textbf{Error Type} & \textbf{Text} & \textbf{GT} & \textbf{Pred}  \\
\midrule
 ECR &{The game} retains some common elements from previous \textbf{Zelda}$_{\mathrm{<MISC>}}$ installments, such as the presence of \textbf{Gorons}$_{\mathrm{<PER>}}$, while introducing \textbf{Kinstones} $_{\mathrm{<PER>}}$ and other new gameplay features. & The Legend of Zelda $_{\mathrm{<MISC>}}$, Gorons $_{\mathrm{<PER>}}$ : characters  & N/A \\
\midrule
 Multi-hop reasoning &\textbf{Parvathy}$_{\mathrm{<PER>}}$ married film actor \textbf{Jayaram} $_{\mathrm{<PER>}}$ who was her co-star in many films on... She has \textbf{two} $_{\mathrm{<NUM>}}$ children, \textbf{Kalidas Jayaram}$_{\mathrm{<PER>}}$ and \textbf{Malavika Jayaram}$_{\mathrm{<PER>}}$.& {Jayaram}$_{\mathrm{<PER>}}$, {Kalidas Jayaram} $_{\mathrm{<PER>}}$: child & N/A\\
\midrule
 Commonsense &\textbf{Olympic Gold} $_{\mathrm{<MISC>}}$ is the official video game of the \textbf{XXV Olympic Summer Games} $_{\mathrm{<MISC>}}$, hosted by \textbf{Barcelona} $_{\mathrm{<LOC>}}$, \textbf{Spain} $_{\mathrm{<LOC>}}$ in \textbf{1992} $_{\mathrm{<TIME>}}$.&XXV Olympic Summer Games $_{\mathrm{<MISC>}}$ , Spain $_{\mathrm{<LOC>}}$ : country & N/A\\

\bottomrule
\end{tabular}}
\caption{
Common errors of RE models
}
\label{tab:re_errors}
\end{table*}

In addition to shared error types with Doc-RE,
we observe two more types of errors based on the WikiEvents and ChFinAnn datasets. 
\paragraph{Multi-events error} In Doc-EE tasks, documents contain multiple events that overlap or occur simultaneously, which requires the model to have sufficient training or advanced techniques to learn the inherent complexity of multi-event documents. In an event-trigger-annotated dataset such as WikiEvents, the model can fail at assigning arguments to the correct events or matching roles to arguments. 
In a trigger-not-annotated dataset like ChFinAnn, event detection errors may occur when models try to identify and differentiate distinct events within the document due to the complex contextual structure of each event. 
\paragraph{Other errors} 
Models face other error types which are mainly associated with previous tasks like entity recognition or caused by the different linguistic features and complexities of datasets. For example, nominal mention recognition and argument span mismatch errors are common in many works, particularly in generative methods.

\paragraph{Noisy data} This issue comprises natural language noises and labeling noises. Real-world documents contain noisy, unstructured, or poorly formatted content, causing difficulties in identifying entities and extracting relations. Natural language can be ambiguous or vague, leading to uncertainty in model inference. 
To overcome the limitations of the cost of creating annotated datasets, researchers commonly apply automatic labeling strategies like distant supervision to generate large-scale training data. However, this leads to several minor problems due to noise and bias: nested entities (i.e., some entities can be embedded within other entities), false negative labels (i.e., entity pairs not known to be related but getting labeled as such in the dataset), and missing ground truth labels. 


Note that Doc-EE errors vary between ChFinAnn and WikiEvents. There could be a number of factors behind the different Doc-EE error distribution between ChFinAnn and WikiEvents. One crucial factor is the diversity in underlying statistics between datasets due to their distinct domains and languages. Compared to the news dataset WikiEvents, the Chinese financial dataset ChFinAnn requires less commonsense comprehension. Each dataset contains unique linguistic features and complexities. WikiEvents has annotated trigger words, and arguments tend to be near the trigger words, whereas ChFinAnn can have events spread across the entire document and is more likely to interfere with other events. Therefore, long-span and multi-events are major error types in ChFinAnn. Moreover, various model designs and approaches usually aim to address specific challenges and optimize performance on the respective dataset.

%% file: 7_challenges.tex
\section{Remaining Challenges}
\label{sec:challenges}

Current difficulties can be broadly categorized into three areas. First, a lot of information is spread out over several sentences. Second, there might be several mentions pointing to the same entity throughout the entire document. Finally, some relations must be deduced from several sentences in order to be discovered. The first two issues have been addressed by existing approaches using attention mechanisms and graph construction, though multiple-step reasoning techniques are less widely used. Progressively, more methods try to use evidence sentences or evidence paths to infer complicated relations. Nevertheless, models continue to struggle with capturing common sense and knowledge-based reasoning as it is difficult to identify a pattern that is extremely similar in the training set or even during pre-training. 
Additionally, creating annotated datasets for this task is time-consuming and expensive, which limits the amount of data available for training and evaluation. Domain-specific datasets differ from main general datasets, but are necessary for identifying relations that are specific to certain domains, understanding domain-specific terminology, and handling the high variability of language used in different domains. There are several promising future directions. First, it is beneficial to incorporate entity coreference systems into doc-IE models, which we believe will play an important role in resolving ECR and multi-hop reasoning errors. Second, more investigations are needed to design a model with multi-hop reasoning capability. Finally, doc-EE and doc-RE can be supplementary tasks to each other. The information produced by these two tasks can provide a more complete picture of the information given in the document.

\section*{Limitations}

A thorough error analysis is conducted with current state-of-the-art algorithms and limitations in existing approaches as well as the remaining challenges are identified for the task of document-level IE. 
However, how a system can effectively address the challenges takes appropriate action, while we exclusively analyze existing studies mainly focused on news, financial, biomedical, and Wikipedia datasets in English and Chinese languages, we acknowledge that the challenges and conclusions drawn may not be generalizable to other domains, languages, or new datasets.

%% file: appendix.tex
\newpage
\section{Statistics of errors for different approaches and datasets}
We evaluated three Doc-RE approaches on the DocRED and Re-DocRED datasets, two Doc-EE methods on the WikiEvents and ChFinAnn datasets. The three Doc-RE approaches represent three state-of-the-art methods for the path-based, graph-based, and sequence-based approaches, while the two Doc-EE studies represent the state-of-the-art of graph-based and generative-based approaches. For each work, we randomly selected 50 errors as provided in Tables 4 and 5, (150 errors in Figure 2, 100 errors in Figures 3 and 4) and examined the causes of them. 
\section{Performance of Existing Methods}
Performance of Doc-RE Existing Methods are shown in Table \ref{tab:Doc-RE DocRED rank}, Table \ref{tab:Doc-RE GDA rank}, and Table \ref{tab:Doc-RE CDR rank}.
Performance of Doc-EE Existing Methods are shown in Table \ref{tab:Doc-EE WikiEvent rank} and Table \ref{tab:Doc-EE ChFinAnn rank}.

\begin{table}[ht]
\centering\small
\begin{tabular}{ll}
\toprule
\textbf{Model} & \textbf{F1}  \\
\midrule
SAIS$^O_{RE+CR+ET}$-SciBERT \cite{xiao_sais_2022} &  87.10  \\
DocuNet-SciBERT-base \cite{zhang_document-level_2021} & 85.30\\
Eider(Rule)-SciBERT-base \cite{xie_eider_2022} & 84.54\\
ATLOP-SciBERT-base \cite{zhou_document-level_2021}  & 83.90	\\
SSAN-SciBERT \cite{xu_entity_2021}  & 83.70\\
\bottomrule
\end{tabular}
\caption{
Doc-RE GDA rank
}
\label{tab:Doc-RE GDA rank}
\end{table}

\begin{table}[ht]
\centering\small
\begin{tabular}{ll}
\toprule
\textbf{Model} & \textbf{F1}  \\
\midrule
SAIS$^O_{RE+CR+ET}$-SciBERT \cite{xiao_sais_2022} & 79.00  \\
DocuNet-SciBERT-base \cite{zhang_document-level_2021} & 76.30\\
Eider(Rule)-SciBERT-base \cite{xie_eider_2022}& 70.63\\
ATLOP-SciBERT-base \cite{zhou_document-level_2021}  & 69.40	\\
SSAN-SciBERT \cite{xu_entity_2021} & 68.70\\
\bottomrule
\end{tabular}
\caption{
Doc-RE CDR rank
}
\label{tab:Doc-RE CDR rank}
\end{table}

\begin{table}[ht]
\centering\small
\begin{tabular}{ll}
\toprule
\textbf{Model} & \textbf{F1}  \\
\midrule
ReDEE \cite{liang_raat_2022} & 81.90  \\
Git \cite{xu_document-level_2021}  & 80.30\\
PTPCG \cite{zhu_efficient_2022}  & 79.40\\
SCDEE \cite{huang_exploring_2021}  & 78.90\\
DE-PPN \cite{yang_document-level_2021} & 77.90\\
HRE \cite{cui_document-level_2022}  & 76.80\\
Doc2EDAG \cite{zheng2019doc2edag}  & 76.30\\
\bottomrule
\end{tabular}
\caption{
Doc-EE ChFinAnn rank
}
\label{tab:Doc-EE ChFinAnn rank}
\end{table}

\begin{table*}[ht]
\centering
\small
\begin{tabular}{lll}
\toprule
\textbf{Model} & \textbf{F1} & \textbf{Ign-F1} \\
\midrule
KD-Rb-l \cite{tan_document-level_2022}  &67.28 & 65.24 \\
SSAN-RoBERTa-large+Adaptation \cite{xu_entity_2021}  & 65.92 & 63.78 \\
SAIS-RoBERTa-large \cite{xiao_sais_2022}  & 65.11 & 63.44\\
Eider-RoBERTa-large \cite{xie_eider_2022}  & 64.79 & 62.85\\
DocuNet-RoBERTa-large \cite{zhang_document-level_2021}  & 64.55	 & 62.40\\
ATLOP-RoBERTa-large \cite{zhou_document-level_2021}  & 63.40	 & 61.39\\
\bottomrule
\end{tabular}
\caption{
Doc-RE DocRED rank
}\label{tab:Doc-RE DocRED rank}
\end{table*}

\begin{table*}[ht]
\centering\small
\begin{tabular}{lcccc}
\toprule
 \multirow{2.5}{*}{\textbf{Model}} &  \multicolumn{2}{c}{\textbf{Arg Identification}}  &  \multicolumn{2}{c}{\textbf{Arg Classification}}\\
 \cmidrule{2-5}
        &  Head F1 & Coref F1 & Head F1 & Coref F1 \\
\midrule
TSAR$_{large}$~\cite{xu_two-stream_2022} & 76.62 & 75.52 & 69.70 & 68.79  \\
EA$^{2}$E~\cite{zeng_ea2e_2022}  & 74.62 & 75.77 &  68.61 & 69.70  \\
BART-Gen\cite{li_document-level_2021} &  71.75 &72.29 &64.57 &65.11  \\
OneIE\cite{li_document-level_2021} &  61.88 & 63.63 & 57.61 &  59.17  \\
BERT-QA\cite{du_event_2020} &  61.05 & 64.59 & 56.16 & 59.36  \\
\bottomrule
\end{tabular}
\caption{
Doc-EE WikiEvent rank
}
\label{tab:Doc-EE WikiEvent rank}
\end{table*}